\documentclass[manuscript,screen]{acmart}

\AtBeginDocument{%
 \providecommand\BibTeX{{%
 \normalfont B\kern-0.5em{\scshape i\kern-0.25em b}\kern-0.8em\TeX}}}

\setcopyright{acmcopyright}
\copyrightyear{2022}
\acmYear{2022}
\acmDOI{XXXXXXX.XXXXXXX}

\usepackage{pifont}
\newcommand{\cmark}{\ding{51}}%
\newcommand{\xmark}{\ding{55}}%
\newcommand{\tabincell}[2]{
\begin{tabular}{@{}#1@{}}#2\end{tabular}
}

\usepackage{graphicx}
\usepackage{amsthm}
\usepackage{amsmath}
\usepackage[noabbrev]{cleveref}
\usepackage{enumitem}
\usepackage{lineno}

\theoremstyle{definition}

\usepackage{makecell}
\usepackage{subcaption,booktabs}
\DeclareMathOperator*{\argmax}{argmax}
\usepackage{colortbl}
\usepackage{tikz}
\definecolor[named]{Gray}{gray}{0.9}
\begin{document}

\title{A Survey On Few-shot Knowledge Graph Completion with Structural and Commonsense Knowledge}

\author{Haodi Ma}
\email{ma.haodi@ufl.edu}
\affiliation{%
 \institution{University of Florida}
 \city{Gainesville}
 \state{Florida}
 \country{USA}
}


\begin{abstract}
 Knowledge graphs (KG) have served as the key component of various natural language processing applications. Commonsense knowledge graphs (CKG) are a special type of KG, where entities and relations are composed of free-form text. However, previous works in KG completion and CKG completion suffer from long-tail relations and newly-added relations which do not have many know triples for training. In light of this, few-shot KG completion (FKGC), which requires the strengths of graph representation learning and few-shot learning, has been proposed to challenge the problem of limited annotated data. In this paper, we comprehensively survey previous attempts on such tasks in the form of a series of methods and applications. Specifically, we first introduce FKGC challenges, commonly used KGs, and CKGs. Then we systematically categorize and summarize existing works in terms of the type of KGs and the methods. Finally, we present applications of FKGC models on prediction tasks in different areas and share our thoughts on future research directions of FKGC. 
\end{abstract}



\maketitle

\section{Introduction}
Knowledge graphs (KGs) are a collection of triples, where each triple represents a relation \textit{r} between the head entity \textit{h} and tail entity \textit{t}. Examples of real-world KGs include Freebase~\cite{bollacker2008freebase}, Yago~\cite{suchanek2008yago} and NELL~\cite{carlson2010toward}. These KGs contain millions of facts and are the fundamental basis for applications like question-answering, recommender systems, and natural language processing. 

Although an immense amount of information is stored in today's large-scale KGs, they are highly incomplete, which makes Knowledge Graph Completion (KGC) a challenge for its downstream applications. Recent trends target learning low-dimension representations of entities and relations for missing link predictions [(Bordes et al., 2013; Trouillon et al., 2016; Dettmers et al., 2017)]. The general idea of these methods is to model and inference various relation patterns between entities based on known facts in the KG. For example, TransE models relations as translation, aiming at inversion and composition patterns. Rotate, as one representative, can infer symmetric, asymmetric, inversion, and composition patterns. 

However, such methods usually require sufficient training triples for all relations to learn embeddings. Previous works~\cite{xiong2018one} show that a large portion of KG relations is long-tail. In other words, they only have a few instances in the KG. For example, about 10\% of relations in Wikidata have no more than 10 triples. Besides, real-world KGs are often dynamic, which means new relations and entities will be added whenever new knowledge is acquired. To tackle these challenges, the model should be capable of predicting new triples given only a small number of examples. 

To address the above challenges, \cite{xiong2018one} proposed two benchmarks, NELL-One and Wiki-One, for few-shot knowledge graph completion (FKGC) and a baseline model called GMatching. The model introduces a local neighbor encoder to learn expensive entity representations with only a few samples for each query relation. One branch of recent works~\cite{zhang2020few, sheng2020adaptive} follows a similar approach and achieves considerable performance by improving the quality of embeddings by considering local graph neighbors. They further argue that entity neighbors should have varied impacts associated with different task relations. Since relations can be polysemous, reference triples should also make different contributions to a particular query. For example, if the task relation is \texttt{isPartOf}, as shown in Figure~\ref{fig:relation example}, such relation has different meanings, e.g., organization-related as in \texttt{(Liverpool, isPartOf, Premier League)} or location-related as in \texttt{(Gainesville, isPartOf, Florida)}. Apparently, for a query \texttt{(Dallas, isPartOf, Taxes)}, the location-related references should be more influential than others. These models~\cite{niu2021relational, sheng2020adaptive} propose to use attention networks to capture the dynamic properties of both entities and references.

\begin{minipage}{0.5\linewidth}
    \begin{figure}[H]
        \centering
        \caption{Example of (a) an entity with diverse roles, credits to ~\cite{sheng2020adaptive} (b)references showing different impact to a particular query}
        \label{fig:relation example}
        \begin{subfigure}{\linewidth}
            \centering
            \includegraphics[width=0.8\linewidth]{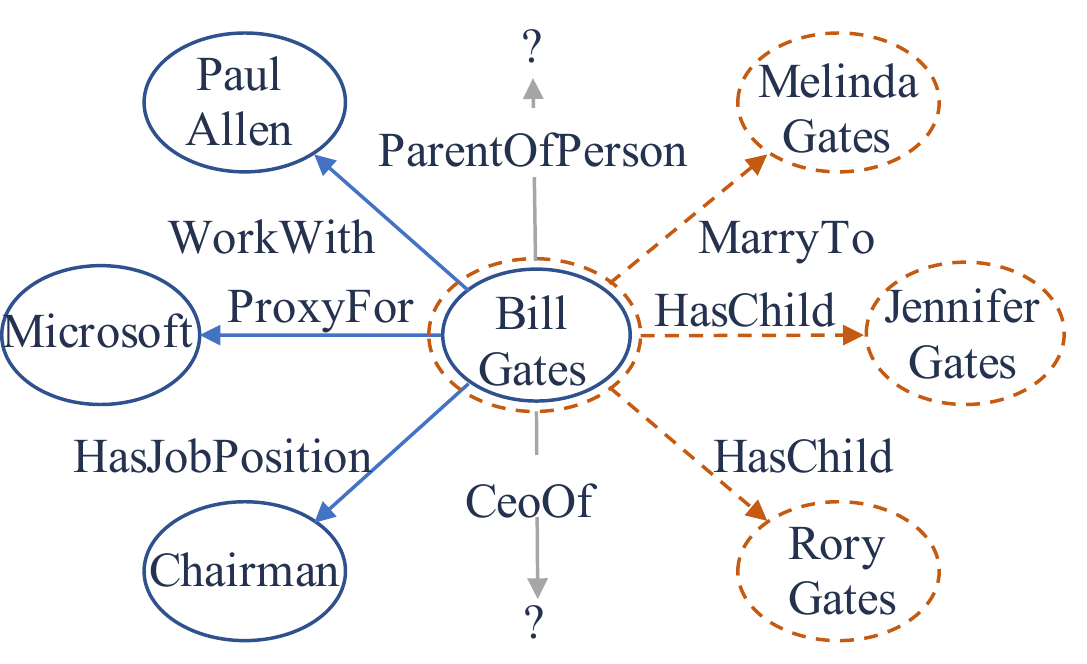}\hfill
            \caption{ }
        \end{subfigure}\par\medskip
        \begin{subfigure}{\linewidth}
            \centering
            \includegraphics[width=0.8\linewidth]{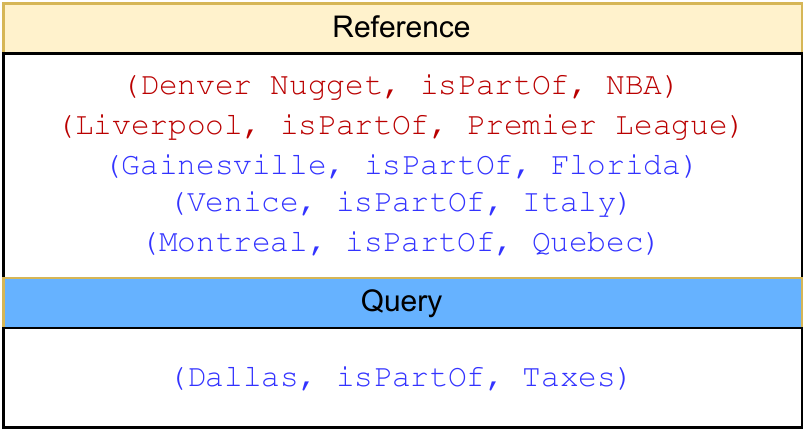}\hfill
            \caption{ }
        \end{subfigure}
        \label{fig:rule-examples-all}
    \end{figure}
\end{minipage}
\begin{minipage}{0.5\linewidth}
    \begin{figure}[H]
        \centering
        \caption{A subset of $\rm ATOMIC_{20}^{20}$, credits to ~\cite{hwang2021comet}}
        \label{ref:fig-commonsense KG example}
        \begin{subfigure}{\linewidth}
        \includegraphics[width=0.9\linewidth]{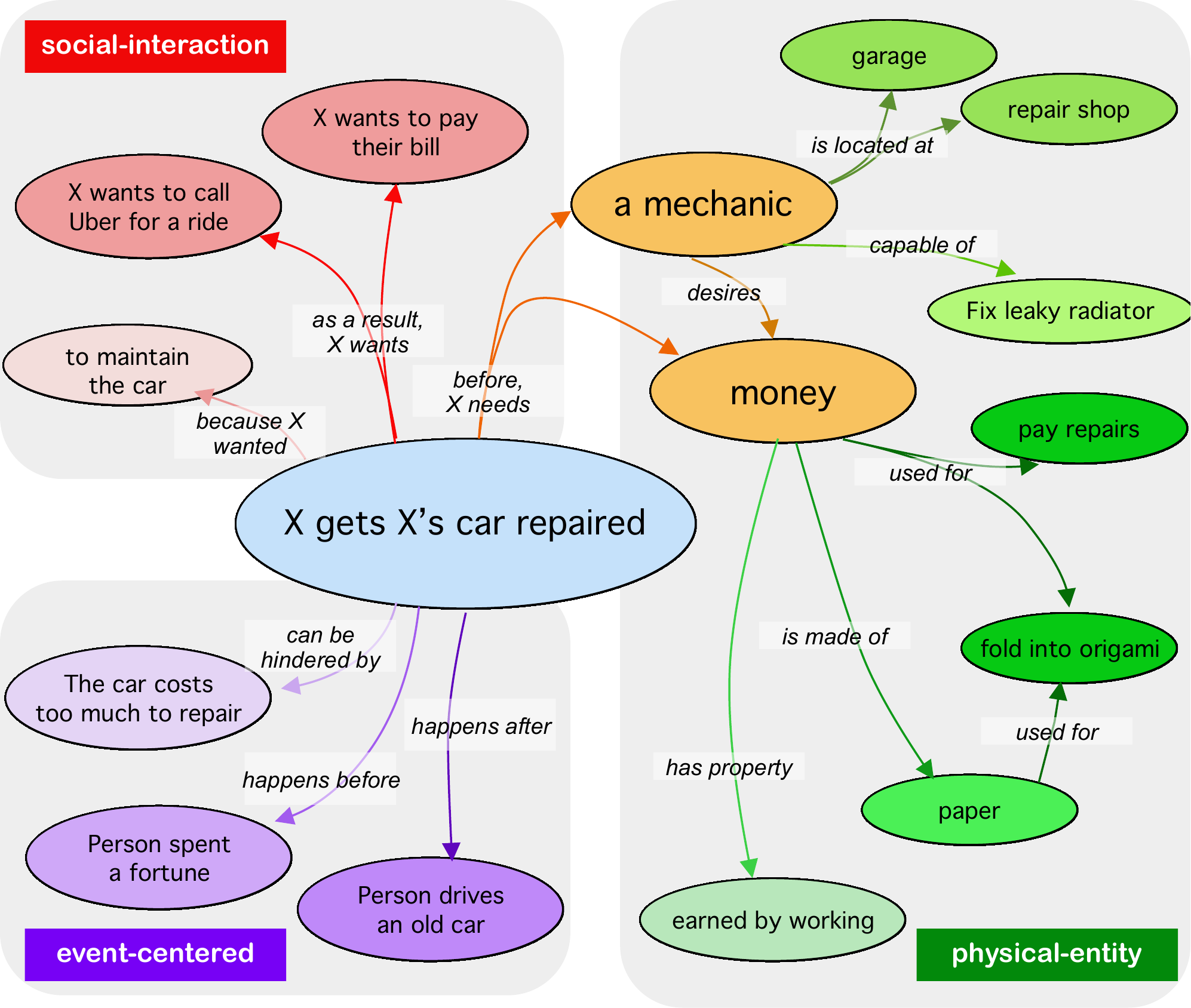}\hfill
    \end{subfigure}
\end{figure}
\end{minipage}

Another track of FKGC models~\cite{chen2019meta, guo2021few, lv2019adapting, zhou2019meta} is developed based on model-agnostic meta-learning (MAML)~\cite{finn2017model}. These models leverage meta-learning to learn the learning process of expressive embeddings of entities and relations with only a few instances. In particular, they use the high-frequency relations in the training set to capture meta-information, which includes common features across different task relations. With a good parameter initialization provided by the meta-information, these models can rapidly adapt to the test tasks where only a few instances are provided for each task relation. 

On the other hand, as a particular type of knowledge graphs, commonsense knowledge graphs (CKGs) like ATOMIC~\cite{sap2019atomic} and ConceptNet~\cite{speer2013conceptnet}, where entities and relations are composed of free-form text, gain less attention from embedding-based models. CKGs are dynamic since entities with unseen text are constantly introduced, which makes them natural benchmarks for FKGC. Besides, entities and attributes in CKGs are usually free-form texts. As shown in Figure~\ref{ref:fig-commonsense KG example}, unlike general KGs that have structured entity and relation names, entity descriptions in CKGs have rich semantic meaning, and implicit semantic relations can be used to infer commonsense knowledge directly, but the such feature also makes CKGs more sparse comparing with general KGs since entities referring to the same concept can be distinct nodes. As shown in~\cite{wang2021inductive}, the average in-degree of ConceptNet and ATOMIC is only 1/15 and 1/8 compared with FB15K-237. Since CKGs do not cleanly fit into a schema comparing two entities with a relation, embedding-based methods are limited to capturing implicit commonsense knowledge. 


Meanwhile, recent progress in training transformer-based contextual language models~\cite{devlin2018bert, radford2018improving} has inspired the interest in using the language models (LMs) as knowledge bases. For example, recent works have focused on querying the LMs with prompts (e.g., \texttt{"Beatles was formed in \_\_"}). COMET~\cite{bosselut2019comet} is a transformer-based KG completion model trained to predict the unseen tail entity conditioning on the head entity and relation on ATOMIC~\cite{sap2019atomic}. BertNet~\cite{hao2022bertnet} takes a step further to directly extract triples for unseen entities from pre-trained language models by automatically paraphrasing an initial prompt for FKGC/KGC tasks. 

Finally, in this survey, we cover typical applications of FKGC models in data science, visual extraction, and medical communities. We further discuss future research directions for FKGC on general and commonsense knowledge graphs based on the observed weakness of current models. 

\section{Preliminaries}
In this section, we first review different KGs. Then we formally define knowledge graph completion and few-shot knowledge graph completion. In the final part of this section, we briefly introduce few-shot learning and meta-learning, which are widely used in FKGC tasks. 
\subsection{Knowledge Graph} \label{sec:KG}
Let $\mathcal{E}$ and $\mathcal{R}$ denote the set of entities and relations, a knowledge graph $\mathcal{G} = \{(e_i, r_k, e_j)\} \subset \mathcal{E} \times \mathcal{R}\times \mathcal{E}$ is a collection of factual triples, where $\mathcal{E}$ represents the set of entities, $\mathcal{R}$ is the set of relations; $e_i$ and $r_k$ are the $i$-th entity and $k$-th relation, respectively. We usually refer $e_i$ and $e_j$ as the head and tail entity. 
A knowledge graph can also be represented as $\mathcal{X} \in \{0, 1\}^{|\mathcal{E}| \times |\mathcal{R}| \times |\mathcal{E}|}$, which is called the adjacancy tensor of $\mathcal{G}$. The $(i, j, k)$ entry $\mathcal{X}_{i, k, j} = 1$ when triple $(e_i, r_k, e_j)$ is true, otherwise $\mathcal{X}_{i, k, j} = 0$. 
A list of commonly-used KGs with their source, size and examples are shown in Table~\ref{tab:survey of KGs}
\subsubsection{structural Knowledge Graph}\label{sec:structural KG}\hfill\\
As introduced earlier, previous works tend to extract semi-structured text to construct knowledge graphs ~\cite{bollacker2008freebase, suchanek2008yago, mahdisoltani2014yago3}. These KGs are usually constructed by crowdsourcing or extracted from crowdsources. 
\\
\\ \noindent\textbf{FreeBase. }
Freebase is a crowdsourced curated KG first introduced in 2008~\cite{bollacker2008freebase} and has been used as a standard baseline KG for many tasks, including KG completion. The most up-to-date and complete version of Freebase contains about 3 billion total triples and about 50 million entities \footnote{https://developers.google.com/freebase/}. A wide-used subset of Freebase, FB15K-237, excludes inverse relations from Freebase and includes 14541 entities, 237 relations, and 272,155 training triples. Relations contained in Freebase are hierarchical, which form a well-defined space of entities and relations that motivates the thread of embedding models.
\\
\\ \noindent\textbf{Wikidata. }
Wikidata is also a crowdsourced KG, containing approximately 78 million data items, with about 23,000 types and 1,600 relations~\cite{paulheim2017knowledge}. At its inception, it was designed to be an alternative method to manage the information found in Wikipedia. As well as providing factual information, Wikidata gives the context around a fact by storing its source. As of 2014, Wikidata supported 287 languages~\cite{vrandevcic2014wikidata}. In 2014 Google transferred the data stored in Freebase into Wikidata~\cite{pellissier2016freebase}. Entities and relations in Wikidata are described through property-value pairs; 
\\
\\ \noindent\textbf{YAGO. }
YAGO is a large knowledge base that is built automatically from Wikipedia. The knowledge graph combines information from Wikipedia in 10 different languages into a whole to provide a multilingual dimension of the knowledge. It also attaches spatial and temporal information to many facts and thus allows the user to
query the data over space and time. As constructed from Wikipedia, YAGO inherits the hierarchy from Wikipedia and uses structural text for entities and relations as well. There exists multiple iterative version of YAGO, including YAGO2 and YAGO3. YAGO3 contains 87 million facts, 10.8 million entities, and 76 million keywords. 

\subsubsection{Commonsense Knowledge Graph}\label{sec:Commonsense KG}\hfill\\
Commonsense knowledge graphs mean to organize commonsense or domain-specific knowledge for downstream applications. Though existing CKGs~\cite{sap2019atomic, speer2013conceptnet} are also commonly constructed by human crowdsourcing, they use free-form text for entities. 
\\
\\ \noindent\textbf{ATOMIC. }
The ATOMIC dataset\footnote{https://homes.cs.washington.edu/~msap/atomic/}, released by \cite{sap2019atomic}, contains 877K tuples covering a variety of commonsense social knowledge around specific event prompts (e.g., "X goes to the store"). ATOMIC contains everyday commonsense knowledge entities organized as if-then relations. It contains over 300K entities in total, and entities are composed of text descriptions with an average of 4.4 words. Specifically, ATOMIC distills its commonsense in 9 dimensions, covering the event's causes (e.g., "X needs to drive there"), its effects on the agent (e.g., "to get food") and its effect on other direct (orimplied) participants (e.g., "Others will be fed").
\\
\\ \noindent\textbf{ConceptNet. }
ConceptNet~\cite{speer2013conceptnet} is a multilingual knowledge graph that connects words and phrases of natural language with labeled edges. Its knowledge is collected from many sources that include expert-created resources, crowdsourcing, and games with a purpose. It represents the general knowledge involved in understanding language using words and phrases of different languages. Such "concepts" can help natural language applications to understand better the meanings behind the words people use. ConceptNet contains over 13 million links between these concepts. 
\\
\\ \noindent\textbf{Visual Genome. }
Instead of just using natural language sources, Visual Genome~\cite{krishna2017visual} collects commonsense knowledge from images as well. It collects dense annotations of objects, attributes, and relations with each image to construct the knowledge. Specifically, Visual Genome contains over 100K images in total, with each image having, on average, 21 objects, 18 attributes, and 18 relations between objects. Since the objects, attributes, and relations are extracted from images, the dataset categorizes them with WordNet~\cite{miller1995wordnet} synsets. In this section, we first review different KGs. Then we formally define knowledge graph completion and few-shot knowledge graph completion. In the final part of this section, we briefly introduce few-shot learning and meta-learning, which are widely used in FKGC tasks. 

\subsection{Few-shot Knowledge Graph Completion}\label{sec:FKGC}
\subsubsection{Knowledge Graph Completion} \label{sec:KGC}\hfill\\
The objective of knowledge graph completion (KGC) is to predict valid but unobserved triples in $\mathcal{G}$. Formally, given a head entity $e_i$ (tail entity $e_j$) with a relation $r_k$, models are expected to find the tail entity $e_j$ (head entity $e_i$) to form the most plausible triple $(e_i, r_k, e_j)$ in $\mathcal{G}$. KGC models usually define a scoring function $f: \mathcal{E} \times \mathcal{R}\times \mathcal{E} \rightarrow \mathbb{R}$ to assign a score $s(e_i, r_k, e_j)$ to each triple $(e_i, r_k, e_j) \in \mathcal{E} \times \mathcal{R}\times \mathcal{E}$ which indicates the plausibility of the triple.

\begin{table*}[!t]
	\centering
	\caption{Survey of existing knowledge graphs and examples.}
	{\footnotesize
	\small
	\begin{tabular} {p{1.6cm} p{5cm} p{2cm} p{7cm}}
		\toprule
		& \bf source & \bf size & \bf examples \\
		\midrule
		\bf Freebase & https://developers.google.com/freebase & 4 relation groups, 2M nodes, 18M edges & \makecell[l]{\texttt{/m/070xg, /sports/sports\_team/colors, /m/01g5v} \\ \texttt{(Seattle seahawks) \-\hspace{72pt} \texttt{(blueish)}}\\ \\ \texttt{/m/06kxk2, /people/person/place\_of\_birth, /m/01\_d4} \\ \texttt{(Carl Foreman) \-\hspace{72pt} (America/Chicago)}}\\ \hline
		\bf Wikidata & https://www.wikidata.org & 1.2k relations, 75M objects, 900M edges & \makecell[l]{\texttt{Austria:Q40, part\_of:P361, European Union:Q458} \\ \\ \texttt{The Beatles:Q1299, location\_of\_formation:P740,} \\ \-\hspace{139pt} \texttt{Liverpool:Q24826} } \\ \hline
		\bf YAGO & https://yago-knowledge.org/ & 87 million facts, 10.8 million entities, and 76 million keywords & \tabincell{l}{\texttt{pl/Henryk\_Pietras, wasBornIn, de/Debiensko} \\ \\ \texttt{fr/Chateau\_de\_Montcony, isLocatedIn, Burgundy}}\\ \hline
		\bf Concept Net & https://conceptnet.io/ & 36 relations, 8M nodes, 21M edges & \makecell[l]{\texttt{/c/en/go\_to\_bed, /r/HasPrerequisite,} \\ \-\hspace{115pt} \texttt{/c/en/get\_ready\_for\_bed} \\ \\ \texttt{/c/en/section\_of\_children's\_books, /r/AtLocation,} \\ \-\hspace{145pt} \texttt{/c/en/bookstore} \\ \\ \texttt{/c/pt/atordoaremos/v, /r/FormOf, }\texttt{/c/pt/atordoar}}\\ \hline
		\bf ATOMIC & https://allenai.org/data/atomic-2020 & 9 relations, 300k nodes, 877k edges & \makecell[l]{\texttt{money, is used for, pay repairs} \\ \\ \texttt{money, is made of, paper} \\ \\ \texttt{PersonX accepts the job, xEffect, joyful}} \\ \hline
 \bf Visual Genome & https://visualgenome.org/ & 42k relations, 3.8M nodes, 2.3M edges, 2.8M attributes & \makecell[l]{\texttt{men.n.01, wears.v.01, backpack.n.01} \\ \\ \texttt{chair.n.01, has.v.01, padding.n.01} \\ \\ \texttt{juice bottle.n.01, on.v.01, desk.n.01}} \\ 
		\bottomrule
	\end{tabular}
}
	\label{tab:survey of KGs}
\end{table*}

\subsubsection{Knowledge Graph Embedding}\label{sec:KGE models}\hfill\\
Knowledge graph embedding (KGE) proposes to project entities and relations into a well-defined space that can be modeled with high-dimensional vectors. Knowledge embedding (KGE) models usually associate each entity $e_i$ and relation $r_j$ with vector representations $\textbf{e}_i$, $\textbf{r}_j$ in the embedding space. Then they define a scoring function to model the interactions among entities and relations. 

KGE models can be generally classified into translation and bilinear models. The representative of translation models is TransE~\citep{bordes2013translating}, which models the relations between entities as the difference between their embeddings. This method is effective in inferencing composition, anti-symmetry, and inversion patterns but cannot handle the 1-to-N, N-to-1, and N-N relations. RotatE~\citep{sun2018rotate} models relations as rotations in complex space so that symmetric relations can be captured, but is as limited as TransE otherwise. ComplEx~\citep{trouillon2016complex}, as a representative of bilinear models, introduces a diagonal matrix with complex numbers to capture anti-symmetry. 
Other models, such as BoxE~\citep{abboud2020boxe}, and HAKE~\citep{zhang2020learning}, can express multiple types of relationship patterns with complex KG embeddings. 

\subsubsection{Graph Neural Network Models}\label{sec:GNN models}\hfill\\
The Graph Neural Network (GNN) has gained wide attention on KGC tasks in recent years~\cite{wang2021mixed, zhang2022rethinking, yu2021knowledge}. With the high expressiveness of GNNs, these methods have shown promising performance. However, SOTA GNN-based models do not show great advantages compared with KGE models while introducing additional computational complexity~\cite{zhang2022rethinking}. For example, NBFNet~\cite{zhu2021neural} and RED-GNN~\cite{zhang2022knowledge} achieve competitive performance on KGC benchmarks, but the leverage of the Bellman-Ford algorithm which needs to propagate through the whole knowledge graph, which restrict their application on large graphs. 

\subsubsection{Few-shot Learning}\label{sec:FSL models}\hfill\\
Few-shot Learning (FSL)~\cite{wang2020generalizing} focus on learning transferable general prior knowledge from existing tasks for new tasks with limited labeled data. It usually adopts a meta-learning framework~\cite{finn2017model} that treats entire tasks as training examples so that the model can adapt fast to new tasks. Specifically, given a set of tasks $\mathcal{T}$ and their training data, in the meta-training phase, the objective of the model is to learn global parameters $\theta'$ that are effective across all tasks in $\mathcal{T}$:
\begin{align*}
 \theta' = \argmax_{\theta} \sum_{\mathcal{T}_i \sim p(\mathcal{T})} \mathcal{L}(\mathcal{D}_{\mathcal{T}_i}, \theta)
\end{align*}
where $p(\mathcal{T})$ is distribution of tasks; $\mathcal{D}_{\mathcal{T}_i}$ is the training data of task $\mathcal{T_i}$; $\mathcal{L}$ is the loss function of the downstream task. Then in the meta-testing phase, $\theta^*$ is taken as the initialized parameters (prior knowledge) that are quickly adapted on a new task $\mathcal{T}_j$:
\begin{align*}
 \theta^* = \mathcal{L}(\mathcal{D}_{\mathcal{T}_j}, \theta)
\end{align*}
where $\mathcal{T}_j$ only has limited labeled data. Previous FSL methods can be generally categorized into (1) metric-based methods~\cite{vinyals2016matching, snell2017prototypical} that exploit task-specific similarity metrics to generalize from support set data to query data; (2) optimization-based methods~\cite{finn2017model, finn2018probabilistic, ravi2016optimization} that aim to find model parameters that are sensitive to changes in the task so that the base learner can quickly adapt to new few-shot tasks with a small number of gradient updates.

\subsubsection{Few-shot Knowledge Graph Completion}\label{sec:FKGC definition}\hfill\\
Following the definition of KGC and FSL, we now formally define few-shot knowledge graph completion (FKGC). \\
Consider a knowledge graph $\mathcal{G} = \{(h, r, t)\} \subset \mathcal{E} \times \mathcal{R}\times \mathcal{E}$ is a collection of factual triples, where $\mathcal{E}$ represents the set of entities, $\mathcal{R}$ is the set of relations, respectively. Given a relation $r \in \mathcal{R}$ and its supporting set $\mathcal{S}_r = \{(h_k, t_k)|(h_k, r, t_k) \in \mathcal{T}\}$, the task is to complete triple $(h, r, t)$ with the tail entity $t \in \mathcal{E}$ missing. In other words, the model needs to predict $t$ from a candidate entity set $\mathcal{C}$ given $(h, r)$. When $|\mathcal{S}_r| = K$ and $K$ is very small, the task is called $K$-shot KG completion. An extreme scenario is when $k = 0$, which means there are no supporting triples. Such a task is also referred to as inductive KGC, zero-shot KGC, or out-of-graph KGC, where models are expected to predict correct relations for unseen entities. 

A few-shot KGC model aims to rank the true entity higher than the the false candidate entities. In FKGC, each training task corresponds to a relation $r \in \mathcal{R}$ with its own supporting/query entity pairs, i.e., $\mathcal{T}_r = \{\mathcal{S}_r, \mathcal{Q}_r\}$. As previously defined, $\mathcal{S}$ contains $K$-shot supporting entity pairs. $\mathcal{Q}_r = \{(h_m, t_m)/\mathcal{C}_{h_m, r}\}$ consists of all queries and the corresponding candidates $\mathcal{C}_{h_m, r}$ which are selected based on the entity type constraint~\cite{xiong2018one}. We further denote all the tasks in training as the meta-training set $\mathcal{T}_{meta-training}$. 

After training on the meta-training set, the few-shot learning model will be tested by predicting facts of new relations $r' \in \mathcal{R}'$. The relations for testing are \textbf{unseen} from the meta-training set, i.e., $\mathcal{R} \cup \mathcal{R}' = \varnothing$. Each relation in the testing phase also has its few-shot supporting and query set: $\mathcal{T}_{r'} = \{\mathcal{S}_{r'}, \mathcal{Q}_{r'}\}$, defined similarly as in meta-training. We denote all tasks in testing as the meta-testing set $\mathcal{T}_{meta-testing}$. The model also has access to a background KG $\mathcal{G}'$ which is a subset of $\mathcal{G}$ with all the relations instead of those in $\mathcal{T}_{meta-training}$ and $\mathcal{T}_{meta-testing}$.

\section{FKGC Models}\label{sec:FKGC-model}
Generally, FKGC models with Structural Knowledge combine KGC models with few-shot learning for various applications. Besides KGE models, GNN-based methods have also shown competitive performance in FKGC since only limited labeled data are provided in the supporting set for each few-shot task. The models that leverage semantic features, on the other hand, utilize prompts to combine structural and semantic information. 

Three main challenges exist for FKGC task~\cite{jiang2021metap}:
\begin{itemize}
    \item (1) \textbf{How to learn the most representative information of triples in the few-shot setting?} General machine learning algorithms require a large number of data for model training, while there are only a few references in the few-shot scenario. Learning representative patterns of different relations from limited triples becomes the key to solving the FKGC problem. 
    \item (2) \textbf{How to decrease the over-reliance on background KGs?} Most prior few-shot methods rely on a background KG to access information from neighborhoods of entities or pre-train the entity embeddings. Some recent models argue that a thorough background KG is not always accessible, and storing it in memory is also space-consuming.
    \item (3) \textbf{How to utilize the negative samples to enhance the model efficacy?} The most intuitive matching approaches generally compare the similarity between queries and positive references while neglecting the similarity between queries and negative references, which can improve the accuracy of triplet validity measurement.
\end{itemize}

In this section, we systematically categorize recent FKGC models with structural knowledge into metric-based methods and optimization-based methods depending on how they adopt FSL techniques and how they tackle the three questions above. Then we go from prompt-based structural models to the ones that take advantage of pretrained language models. A list of representative FKGC works with their open-source dataset/codes is provided in Table\ref{tab:representative works}

\subsection{Metric-based Method}
Existing metric-based FKGC models share the framework of either Matching Network~\cite{vinyals2016matching} or Translation Network~\cite{bordes2013translating}. 

For models that are built based on Matching Network, they first implement a GNN-based entity encoder to generate entity embeddings. Then an aggregation module is applied on entity pairs in the supporting set to compute the embedding of each relation. Finally, the model computes the probability of acceptance of each query triple based on its similarity to supporting triples. KGE models like TransE~\cite{bordes2013translating} and ConvE~\cite{dettmers2018convolutional} are also widely used in entity encoder as the intermediate representation to be further enhanced with other information. 

Following this framework, GMatching~\cite{xiong2018one} is the first work to solve the one-shot KGC problem. It first proposes a neighbor encoder, which utilizes the local graph structure to generate better entity embedding. The motivation here is that, although the entity embedding of previous KGE models can encode relational information, previous work~\cite{xiong2017deeppath} shows that explicitly modeling structure patterns like path can still benefit relational prediction. The neighbor encoder in GMatching encodes only the one-hop neighbors of each given entity, i.e., a set of (relation, entity) tuples to guarantee it is general to large-scale KGs. Specifically, starting with pre-trained KGE embeddings for every tuple in the one-hop neighbor set, GMatching applies a feed-forward layer to encode the interaction between the relation and the entity in each tuple. A neighbor encoder is then applied on both supporting entity pairs and query entity pairs to generate each representation. Then, the model exploits an LSTM-based recurrent processing block~\cite{vinyals2016matching} to perform multi-step matching between the reference pair and each query pair. The matching scores are finally used to rank every entity in the candidate set of each query. Besides proposing the first baseline model on FKGC task, the work also proposes two widely-used benchmarks: NELL-One and Wiki-One~\cite{xiong2018one}. Both are built following the FKGC task setting. More statistics and details are provided in Table~\ref{tab:FKGC benchmarks} and Sec~\ref{sec:resources}.

\begin{table*}[t]
\centering
\resizebox{1\textwidth}{!}{
\begin{tabular} {c|c|c|c|c}
\toprule
Method & Learning Task & FSL technique & Venue & Code/Data Link \\
\midrule
\rowcolor{Gray}
\multicolumn{5}{c}{Structural FKGC}\\
\midrule
GMatching~\cite{xiong2018one} & Relation prediction &  Matching-based & EMNLP'18 & \url{https://github.com/xwhan/One-shot-Relational-Learning}\\
\midrule
FSRL~\cite{zhang2020few} & Relation prediction & Matching-based & AAAI'20 &  \url{https://github.com/chuxuzhang/AAAI2020_FSRL}\\
\midrule
FAAN~\cite{sheng2020adaptive} & Relation prediction & Matching-based & EMNLP'20 & \url{https://github.com/JiaweiSheng/FAAN}\\
\midrule
GEN~\cite{baek2020learning} & Relation prediction & Matching-based & NeurIPS'20 & \url{ https://github.com/JinheonBaek/GEN}\\
\midrule
REFORM~\cite{wang2021reform} & Relation prediction &  Matching-based & CIKM'21 & \url{https://github.com/SongW-SW/REFORM}\\
\midrule
MetaR~\cite{chen2019meta} & Relation prediction & Matching-based & EMNLP'19 & \url{https://github.com/AnselCmy/MetaR}\\
\midrule
GANA~\cite{niu2021relational} & Relation prediction & Matching-based & SIGIR'21 & \url{https://github.com/ngl567/GANA-FewShotKGC}\\
\midrule
MetaP~\cite{jiang2021metap} & Multi-hop relation prediction & Metric-based & SIGIR'21 &  \url{https://github.com/jzystc/metap}\\
\midrule
Meta-KGR~\cite{lv2019adapting} & Multi-hop relation prediction  &  Optimization-based & EMNLP'19 &  \url{https://github.com/THU-KEG/MetaKGR}\\
\midrule
ADK-KG~\cite{zhang2022hg-adk-kg} & Multi-hop relation prediction & Optimization-based & SDM'22 &  \url{https://github.com/ADK-KG/ADK-KG}\\
\midrule
ZS-GAN~\cite{qin2020generative} &  Multi-hop relation prediction & Optimization-based & NeurIPS'21 & \url{https://github.com/Panda0406/Zero-shot-knowledge-graph-relational-learning}\\
\midrule
\rowcolor{Gray}
\multicolumn{5}{c}{Commonsense FKGC}\\
\midrule
ConMask~\cite{shi2018open} & Relation prediction &  Text fusion-based & AAAI'18 & \url{https://github.com/bxshi/ConMask}\\
\midrule
MIA~\cite{niu2021open} & Relation prediction & Text fusion-based & WWW'21 &  \url{--}\\
\midrule
InductiveE~\cite{sheng2020adaptive} & Entity prediction & LM-based & IJCNN'21 & \url{https://github.com/BinWang28/InductivE}\\
\midrule
BERTRL~\cite{zha2022inductive} & Relation prediction & LM-based & AAAI'22 & \url{ https://github.com/zhw12/BERTRL}\\
\midrule
OntoPrompt~\cite{ye2022ontology} & Entity prediction & Prompt-based & WWW'22 & \url{https://github.com/zjunlp/KnowPrompt}\\
\midrule
ZS-SKA~\cite{chen2019meta} & Relation prediction & Prompt-based & arxiv & \url{--}\\
\midrule
COMET~\cite{bosselut2019comet} & Relation prediction & LM-based & AAAI'21 & \url{https://github.com/atcbosselut/comet-commonsense}\\
\midrule
BERTNet~\cite{hao2022bertnet} & Triple prediction & LM-based & -- &  \url{https://github.com/tanyuqian/knowledge-harvest-from-lms}\\
\toprule
\end{tabular}
}
\caption{Representative FKGC methods with open-source code/data.} 
\label{tab:representative works}
\end{table*}

Sharing the same idea, FSRL~\cite{zhang2020few} extends GMatching to the few-shot setting. It further proposes a relation-aware heterogeneous neighbor encoder to enhance entity embeddings based on the heterogeneous graph structure and attention mechanism so that the model can encode the different impacts of different neighbors on the task relation. The main argument here is that different neighbors should impact the task relation differently, which models like GMatching neglect. For example, taking \texttt{ParentOfPerson} as the task relation, the neighbor \texttt{(MarryTo, Melinda Gates)} should have a higher weight compared with \texttt{(CeoOf, Microsoft)}. To tackle such an issue, FSRL introduces an attention module to generate entity embeddings by assigning different attention weights when encoding all neighbors. 

By applying the attentive neighbor encoder, FSRL acquires the representation of each entity pair in the supporting set. It then implements an RNN-based aggregator to model interactions between supporting entity pairs for each task related to generate an informative representation of the entire supporting set. Inspired by aggregating node embeddings with recurrent neural network~\cite{hamilton2017inductive}, FSRL applies a recurrent autoencoder aggregator on all entity pairs. In order to formulate the embedding of the reference set, it aggregates all hidden states of the encoder and extends them by adding residual connection~\cite{he2016deep}and attention weight.

With the aggregated representation of the reference set, FSRL applies a matching network to discover similar entity pairs of the reference set. Instead of comparing each reference entity pair with the query pair, a similar recurrent matching processor with LSTM cells is used to directly compute the similarity between the reference set and query entity pair for the final answer ranking. During the training session, each time model samples a task relation and optimizes the model for that task. The model will sample few-shot entity pairs as the supporting set and a batch of query entity pairs. Negative training sets are constructed by polluting the tail entities in query entity pairs. Meta-learning is exploited in the gradient descent step for parameter optimization so that FSRL can transform well onto test few-shot relations. 

Although FSRL~\cite{zhang2020few} proposes to treat neighbors differently based on their relevance to the central entity, it still assigns fixed weights to all neighbors throughout all task relations. Such a solution leads to static entity embeddings in different tasks, hurting the system's effectiveness. FAAN~\cite{sheng2020adaptive}, taking a step further, argues that entity neighbors should have varied impacts with different task relations. For example, \texttt{SteveJobs} is associated with task relations \texttt{HasJobPosition} and \texttt{HasChild}. Intuitively, if the task relation is \texttt{CeoOf}, the model should pay more attention to the job position role of entity \texttt{SteveJobs} than the family role. 

Besides, task relations can have different meanings under different contexts. For example, if the task relation is \texttt{isPartOf}, as shown in Figure~\ref{fig:relation example}, such relation has different meanings, e.g. organization-related as in \texttt{(Liverpool, isPartOf, Premier League)} or location-related as in \texttt{(Gainesville, isPartOf, Florida)}. Apparently, for a query \texttt{(Dallas, isPartOf, Taxes)}, the location-related references should be more influential than others. Therefore, the reference (supporting) triples should also contribute variously to different queries. 

To address the above challenges, FAAN proposes an adaptive attentional neighbor encoder to model entity embeddings with one-hop entity neighbors. They also follow TransE~\cite{bordes2013translating} to model the task relation embedding as a translation between the head and tail entity embeddings, i.e., $\textbf{r} \approx \textbf{h} - \textbf{t}$. Then, to further model various roles of the reference entities, FAAN train an attention metric based on the relevance of entity neighbor relations and the task relation to further obtain a role-aware neighbor embedding for each entity in the reference set. The encoder allows dynamic attention scores adaptive to different task relations. The adaptive mechanism helps to capture the diverse roles of entities based on the different contributions of neighbors. The final representation of each entity encodes both the pre-trained embedding and its role-aware neighbor embedding. 

With the enhanced entity representation provided by the encoder, FAAN further applies a stack of Transformer blocks for supporting and query triples to capture various meanings of the task relation. It borrows the idea of learning dynamic KG embeddings from \cite{wang2019coke}. For each element, it passes the element embedding and position embedding through several Transformer blocks to acquire meaningful entity pair embeddings. 

Then, instead of using static representations when predicting different queries, FAAN obtains a general adaptively representation of the supporting set by aggregating all the references with their attention score to the task relation. FAAN also uses meta-training in the same fashion as FSRL, i.e., the model is trained on different task relations in the meta-training set to generate a set of parameters that performs well across all the tasks and can quickly adapt to few-shot tasks in the test set. With all the above, FAAN improves the quality of entities and reference representations by capturing their fine-grained meanings. Sharing a similar matching score as in FSRL, FAAN outperforms previous models on FKGC task. 

HARV, on the other hand, focuses on capturing the differences between neighbor relations and entities and interaction between relations, which are previously neglected. It introduces a hierarchical neighbor aggregator for central entity representation by separating the information between the head entity and relation (relation-level) and between the relation and tail entity (entity-level). The relation-level attention weights are computed based on the head entity and relation embeddings. The relation-level embeddings are generated by aggregating neighbor relations of head entity $h$ with such attention. Concatenations of relation-level embedding and each tail entity are then used to generate the entity-level attention weights. The two-level weights finally generate the triple-level weight, which is used to compute the enhanced entity representation. Interactions between relations are taken into account by the relation encoder. The encoder is an extension of the LSTM aggregator in FSRL with Bi-LSTM, which updates representations of all support entity pairs. The concatenation of the embedding of support entity pairs and the embedding from the Bi-LSTM encoder is used as the final representation of each entity pair, and the supporting set is represented by an attention-based aggregation of all support entity pairs. 

In addition, GEN~\cite{baek2020learning} investigates an out-of-graph FKGC scenario for relation prediction between unseen entities or between seen and unseen entities. It is meta-learned to extrapolate the knowledge from seen to unseen entities, and transfer knowledge from entities with many to few links. GEN further develops a stochastic embedding layer for transductive inference to model uncertainty in the link prediction between unseen entities. Gen is compatible with any GNNs. Specifically, two GENs are employed at the meta-training stage for both inductive and transductive link prediction. The first GEN is inductive GEN. It learns to encode the unseen entities that are not observed and predicts the links between seen and unseen entities. The second GNN, respectively, is transductive GEN. It takes a step further to learn to predict the links between unseen entities themselves. To enable transductive inference, the meta-learning framework in GEN can simulate the unseen entities during meta-training while they are not observed in conventional learning schemes. Also, since link prediction for unseen entities is inherently unreliable, which gets worse in the few-shot scenario where only a few triplets are available for each entity, GEN learns the distribution of unseen representations for stochastic embedding to account for the uncertainty.
Further, we apply a transfer learning strategy to model the long-tail distribution. These lead GEN to represent the unseen entities that are well aligned with the seen entities. As mentioned, a naive GEN may be affected by the intrinsic unreliability of few-shot out-of-graph link prediction due to the uncertainty of unseen entities' representations caused by lacking supporting triples. The stochastic layer tackling this issue embeds an unseen entity by learning the distribution over that entity embedding. GEN also models the source of uncertainty on the output embedding from the transductive GEN with Monte Carlo dropout. 

More recently, REFORM~\cite{wang2021reform} proposes an error-aware module to control the negative impact of errors affecting FKGC. It slightly varies from the original FKGC to predict the missing relation category of the query entity pair from the few-shot relation categories. Since most real-world KGs are automatically constructed, many errors are incorporated into KGs without manual validation. Such errors significantly alleviate the performance of previous methods on FKGC, especially when there are only a few supporting triples to rely on. The neighbor encoder of REFORM focus on selecting the most reliable neighbors with an attention mechanism to enhance entity representations. The attention weight matrix is trained with pre-trained embeddings (in REFORM, TransE) to ensure those correct neighbors have higher weights. The matrix is then normalized using a softmax function to acquire a robust embedding for each entity. Reference entity pairs are represented by the concatenation of their head and tail entity embeddings. Then, to generate robust embedding for relations in the supporting set, REFORM contains a cross-relation aggregation module based on a transformer encoder to capture the relation correlations and support instances. The transformer encoder makes each input embedding participate in the encoding of all other input embeddings based on a multi-head attention mechanism. Then in the error mitigation module, REFORM exploits the graph convolution network (GCN) to generate confidence weights of various relations for each query task. The confidence weights can be considered attention weights to limit errors' impact. Specifically, REFORM builds a query-oriented graph to measure the effect of different supporting instances on a specific query relation. The GCN is trained to minimize the loss that a query relation is grouped into the wrong category. 

A representative that uses Translation Network is MetaR. The idea is that instead of encoding neighbor information, MetaR focus on transferring the common and shared information within one task from reference instances to query triples. Such information is referred to as relation meta in MetaR. The relation-meta learner generates representations of entity pairs from head and tail entity embeddings in the supporting set. Given the head and tail entity pairs in the supporting set, the learner first extracted entity-pair specific relation meta through a fully connected neural network which uses LeakyReLU~\cite{maas2013rectifier} as the activation function. The final relation meta of a task is the average of all entity-pair specific relation meta in the current supporting set. 

MetaR also exploits Meta-learning to accelerate the learning process, which is referred to as gradient meta. As mentioned in Section~\ref{sec:FKGC definition}, the model should be able to update a new few-shot task rapidly. Inheriting the idea of TransE~\cite{bordes2013translating}, MetaR applies a similar score function $||\textbf{h}_i + \mathcal{R}_{\mathcal{T}_r} - \textbf{t}_i||$ to calculate the score of each entity pair with the relation meta. Then, by minimizing the loss over the supporting set with the score of all positive and negative triples, the gradients of parameters can indicate how they should be updated. Following this gradient update rule, MetaR can make quick updates on relation meta and use the updated one to score the query set with the same scoring function. The model is trained to minimize the sum of query loss over all the tasks in one batch. Compared with GMatching~\cite{xiong2018one}, which relies on a background knowledge graph, our MetaR is independent of them, thus, it is more robust as background knowledge graphs might not be available for few-shot link prediction in real scenarios.

GANA~\cite{niu2021relational}, taking a step further, extends MetaR by refining embedding and relation meta computation with attention mechanism and an LSTM aggregator. The motivation here is that noise neighbor information may hurt the model when the neighbors are spare or even if no proper neighbor is available to represent the few-shot relation. GANA proposes a global-local framework. At the global stage, a gated and attentive neighbor aggregator is built to accurately integrate the semantics of a few-shot relation's neighborhood, which helps filter the noise neighbors even if a KG contains extremely sparse neighborhoods. The head and tail entities associated with the few-shot relation and their neighborhoods are combined to eliminate noise neighbor information due to the sparse neighborhood. A gating mechanism could determine the importance of the neighborhood representation to represent a few-shot relation. Specifically, a graph attention network (GAT)-based neighbor encoder is developed at the global stage to capture different impacts of neighbors to improve the quality of entity embedding. The encoder generates the attention weight for each neighbor based on a trainable linear transformation matrix. GANA employs a gate value with linear transformation to eliminate noise neighbors due to the sparse neighborhood to automatically determine the impact of the neighbor of an entity for the few-shot task relation. An entity is then represented by combining its entity embedding with its neighbor representations. The final triple neighbor representation of the supporting set is the concatenation of the head and tail representations. With the supporting set encoded, GANA employs an attentive Bi-LSTM encoder to integrate multiple neighborhood representations of a query relation in the support set. The query relation representation is a weighted sum of the final hidden states of the Bi-LSTM by combining all the neighbor embeddings in the supporting set. For the local stage, a meta-learning-based TransH(MTransH) method is designed to model complex relations and train our model in a few-shot learning fashion. The reason to use TransH is its ability to model complex relations. A similar loss function is applied with MAML approach to learning well-initialized parameters over all few-shot (query) relations in the meta-training set. 

Another similar FKGC model HiRe~\cite{wu2022hierarchical}, can be seen as an extension of GANA. It proposes to jointly capture three levels of relational information: entity-level, triple-level, and context-level. Contrastive learning is used to encode the union of the neighbors of the head and tail entities together in a triple to encode a wider context. HiRe proposes a context encoder for the target triplet to learn the embeddings of its true/false contexts based on the self-attention mechanism so that important neighbors within the context would be given higher weights. Furthermore, a contrastive loss is employed to pull close the triplet towards its actual context and push it apart from its false context. Then at the triple-level relational learning stage, instead of LSTM, HiRe develops a transformer-based meta-relation learner to capture interactions among reference triples and generates meta relational representation of target relations. Finally, HiRe employs a TransD-based~\cite{ji2015knowledge} meta score function to capture the diversity of entities and relations. MAML-based training strategy is also applied similarly to GANA. With the three-level relational information, HiRe performs better on NELL-One and Wiki-One compared with state-of-art models. The ablation study further proved that all three levels of relational information are crucial to the performance of HiRe, which future models can further leverage.

Meta-iKG, another recent work on this track, proposes to utilize local subgraphs to transfer subgraph-specific information and rapidly learn transferable patterns through meta-gradients with meta-learning. Graph neural network is recently incorporated into inductive relation reasoning to capture multi-hop information around the target triplet. For example, GraIL~\cite{teru2020inductive} proposes a subgraph-based relation reasoning framework to process unseen entities. CoMPILE~\cite{mai2021communicative} extends the idea by introducing a node-edge communicative message-passing mechanism to model the directed subgraphs. Meta-iKG can be interpreted as an extension of CoMPILE method to FKGC.
Instead of being limited to transductive settings and unable to process unseen entities, Meta-iKG targets a few-shot inductive KGC task, including new entities in the test set. The model splits relations into few-shot and large-shot relations with a threshold $\mathcal{K}$ on relation instances number and meta-train with large-shot relations to find well0initialized parameters and adapt the model on triples with few-shot relations following the framework of MAML. Inheriting the structure from MetaR, Meta-iKG first extracts direct enclosing subgraphs between target and tail entities at the relation-specific learning stage. Then an inductive node labeling function is applied to identify the different roles of entities in the subgraph. The node embedding is initialized by the distances to the target entities to embed the relative position of each node in the subgraph. Meta-iKG then follows the idea of CoMPILE to use communicative message-passing neural network to score each subgraph to encode its plausibility of the target triple as the task loss. Regular meta-learning steps promise performance on few-shot relations. However, they may introduce bias to the updated parameters since the task relation query set only updates the final parameters. To guarantee the performance of Meta-iKG on large-shot relations as well, it introduces the large-shot relation update procedure, which further updates the final parameters using the support set with a lower learning rate. This operation enables Meta-iKG to generalize well on the whole inductive dataset.

To tackle the KG-dependent problem and further exploit negative samples in the training stage, a meta pattern learning framework, MetaP~\cite{jiang2021metap}, is proposed. Patterns in data are representative regularities to classify data. Triples in KGs also follow relation-specific patterns, which can be used to measure the validity of triples. The pattern of a relation refers to the regularity of feature co-occurrence of the head entity, relation, and tail entity. MetaP designs a pattern learner based on convolutional filters to extract patterns of triples directly. It can learn latent representations of relation-specific patterns from limited references and thus is independent of the background KG. Besides, by leveraging negative references, MetaP can measure the validity of query triples more accurately. A pattern matcher with a validity balance mechanism (VBM) is proposed to predict the probabilities of whether patterns of query triples are positive or negative.

\subsection{Optimization-based Method}
Optimization-based FKGC models rely on MAML for model optimization. In other words, such models tackle the challenge of the few-shot relation prediction problem by optimizing GNN with MAML.

Since methods like GMatching and FSRL only focus on fact prediction and exploit only one-hop neighbors, they miss more structure information provided in KGs, and results lack interpretation. Accordingly, multi-hop relation reasoning was proposed to infer facts using multi-hop reasoning paths. e.g., \texttt{(The Beatles, FoundIn, Liverpool) $\land$ (Liverpool, PartOf, British) $\rightarrow$ \texttt{(The Beatles, BaseIn, British)}}. Recent models~\cite{lin2018multi, das2018go} propose multi-hop reasoning methods, which leverage the symbolic composition information of relations in KGs to achieve explainable reasoning results. These works also state multi-hop reasoning as a sequential decision process and exploit reinforcement learning to tackle such tasks. A meta-based algorithm for multi-hop reasoning (Meta-KGR) sharing similar ideas is then proposed to provide explainable and effective few-shot relations reasoning. Specifically, Meta-KGR introduces a reinforcement learning framework to model the multi-hop reasoning process, where a recurrent neural network encodes the search path. It then adopts MAML to learn effective meta-parameters from high-frequency relations that could quickly adapt to few-shot relations. 

Specifically, deriving the on-policy reinforcement learning (RL) from ~\cite{lin2018multi}, the multi-hop reasoning is treated as a Markov Decision Process (MDP): give the query relation $r_q$, the model starts from the source entity $e_s$, sequentially step through several relations and entities until it arrives at the target entity $e_o$. The MDP module includes (1) state, which is the entity at the current step. All the states share the source entity and task relation as the global context. (2) action: the action space at state $s_t$ includes all the current entity's outgoing relation and entity tuples. (3) reward: the model will receive a terminal reward equal to 1 if it reaches the correct target entity. Otherwise, a reward will be given based on the similarity between the target entity and the current entity using pre-trained KG embeddings. The policy network is then used to determine action at each state. Then,  Meta-KGR applies the policy network considering the search history over background KG. The search history before the current step is encoded with LSTM. The action space is represented by stacking all the action embeddings in the action space at the current step. The policy network is trained to maximize the expected reward over all triples. In meta-learning, Meta-KGR employs a meta-policy network similar to MAML so that Meta-KGR can quickly adapt to a relation-aware policy network for every query relation with well-initialized parameters learned in this stage.

FIRE~\cite{zhang2020few2} extends Meta-KGR with a heterogeneous neighbor aggregator and a search space pruning strategy. Specifically, FIRE leverages on-policy reinforcement learning to model the path of multi-hop reasoning and encodes entity embedding using multi-hop heterogeneous structural information. It then prunes the reasoning search space using knowledge graph embedding to improve the reasoning efficiency. Meta-learning is also applied in the optimization procedure so that the learned parameters can be fast adapted for few-shot task relations. 

Specifically, since the original RL module in Meta-KGR does not encode the heterogeneous structure information into the entity embedding, FIRE keeps the structure encoding module as in FSRL to enhance entity embeddings. Note that some entities in KGs have a large number of neighbors, making the action space redundant at specific steps. Different from ~\cite{das2018go, lin2018multi} that cut edges based on centrality score, FIRE takes structural correlation between states as an important feature to guide action search and applies a knowledge-aware search space pruning strategy. The model keeps only top-$m$ most correlated entities at each step based on the structural correlation between entities at step $t$ and possible candidates at step $t+1$ with pre-trained KGE like TransE. Fast adaptation with meta-learning is then utilized to learn well-initialized parameters that can quickly generalize to few-shot relations. 

More recently, ADK-KG~\cite{zhang2022hg-adk-kg}further improves FIRE by developing a text-enhanced heterogeneous graph neural network to encode node embeddings, where entity and relation embeddings are pre-trained using content information and augmenting MAML with task weight. It is the first work to leverage a pre-trained language model to capture content information in the FKGC task. The reinforcement learning module is similar to the ones in FIRE and Meta-KGR and generates the encoding of each entity. The problem is that RL only encodes the reasoning process but ignores the content and structural information in KG. ADK-KG thus develops a text-enhanced heterogeneous graph neural network to enrich entity embeddings. Firstly, for each entity and relation in KG, ADK-KG extracts their text information as their content features. It then merges all text features of entities and relations and feeds them into the pre-trained BERT language model~\cite{devlin2018bert}to obtain the corresponding content feature vector. For enumerated content (e.g., entity and relation ids in Wikidata), ADK-KG applies one-hot encoding to convert it to a binary feature vector. After that, a neural network is utilized for encoding and aggregating content embeddings of selected neighbors for each relation type. The selected neighbors include first-order neighbors and relations and also high-order neighbors sharing the same relation and the first-order ones from a random walk. Finally, because different relation types of neighbors will make different contributions to the final entity representation, the model employs the attention mechanism to utilize these relation-type-based neighborhood embeddings to generate the final embedding of each entity. Such embeddings are then used in RL-based reasoning to replace pre-trained entity embeddings. In the meta-learning step, since relations in KG usually have different meanings, AKD-KG assign different weights to them with self-attention mechanism. 

Another recent work extending Meta-KGR and FIRE is THML~\cite{zheng2021hardness}. THML argues that RL-based models' generalization is usually limited by low reasoning performance on hard relations (relations with high training loss). THML challenges this problem in FKGC by identifying the hard relations at each training batch and then further training the model on those effectively generated new hardness-aware training batches from both relation and relation cluster levels. THML also formulates the reasoning process as an MDP as in Meta-KGR and FIRE at the hardness-aware meta-reinforcement learning module. The main difference is that to solve the sparse reward caused by false reasoning paths and efficiency concerns, THML splits the reward into three parts. The terminal reward is the same as in FIRE, except that THML uses ConvE~\cite{dettmers2018convolutional} for pre-trained embeddings. The path reward encodes the reasoning chain length to encourage the model to find the target entity with a relation chain that is as short as possible since shorter paths often provide more reliable reasoning than longer paths~\cite{das2018go}. A path may be declined if the length exceeds 3. Another problem with KG reasoning is that models tend to infer paths with similar semantic meaning in the training stage, which may lead the model into a local-optimal path. THML thus proposes that the diversity reward encourages the model to find different paths. Then, instead of random sampling triple queries at the training stage, THML applies a two-level hardness-aware sampling strategy. Relation level hardness-aware sampling ranks the reasoning accuracy of all relations in a batch online to select hard relations for the next batch. At relational-cluster level hardness-aware sampling, THML obtains pre-trained TransE embeddings for all relations, then performs the k-means algorithm to form relation clusters. It then selects the cluster with the hard relation with the lowest accuracy at the current batch as the hard cluster and adds it to the next batch. 

Apart from the methods discussed above, there are also other types of optimization-based solutions on FKGC. For example, ZSGAN~\cite{qin2020generative} studies zero-shot KGC by establishing the connection between text and knowledge graph with generative adversarial networks. The motivation is that the semantic features of new classes can be reflected by their textual descriptions. Moreover, textual descriptions contain rich and unambiguous information, which is critical for large-scale recognition tasks. The core of ZAGAN is the design of a conditional generative model to learn the qualified relation embeddings from raw text descriptions. 

ZSGAN leverages a feature encoder for real data representations to generate reasonable real data distribution from KG embeddings. The feature encoder is trained in advance from the training set and fixed during the adversarial training process. The neighbor encoder only considers one-hop neighbors of each entity and used GCN~\cite{schlichtkrull2018modeling} to generate the structure-based representation of each entity. Then a feed-forward layer is used as the entity encoder to extract the information from each head, tail entity pair. The relation fact representation is finally formulated as the concatenation of head and tail neighbor embeddings and the entity pair embedding. 

Given text representations, the generator is to generate reasonable relation embeddings that capture the corresponding relational semantic information in the knowledge graph feature space. Based on this, the prediction of query relations is converted to a supervised classification task. Specifically, text embedding is the vector sum of word embeddings weighted by TF-IDF values after removing stop-words and punctuations. The text embedding is passed to the generative adversarial model (GAN) to generate the relation embedding. On the contrary, the discriminator seeks to separate the fake data from the real data distribution and identifies the relation type as well. The input features are first transformed via a fully-connected (FC) layer with LeakyReLU~\cite{maas2013rectifier}. Two network branches follow after. The first branch is a FC layer that acts as a binary classifier to separate real data from fake data. The other branch is classification performance. In order to stabilize training behavior and eliminate mode collapse, ZSGAN also adopts the gradient penalty, which penalizes the model if the gradient norm moves away from the target norm value 1. Similarly, RAN~\cite{zhang2020relation} worked out a general feature generation-based framework for addressing unseen relations in both few-shot KG completion with unseen relations and few-shot relation extraction from text.

Unlike previous works that leverage entity pair matching, P-INT~\cite{xu2021p} utilizes the paths from the head to the tail entities to represent an entity pair and computes the interactions between paths for the FKGC problem. The motivation is still to involve more structural and expressive information and exclude noise neighbors in few-shot reasoning. To extract the support subgraph for each support entity pair, P-INT employs a two-side BFS algorithm~\cite{xiong2017deeppath} to prune the search space. The intersected neighbors of the left and right paths are used to generate paths from head to tail entities with different lengths. The relations in these paths are then combined as a set of supporting relations. Pre-trained TransE embeddings are used to compute the similarity between each pair of relations in the KG. Then to reason the query subgraph, P-INT extends the limited number of neighbors with a fixed number of steps and, for each of them, gets the maximum similarity with the supporting relations and returns top-$L$ ones. After $T$ hops,  the model extends a query subgraph with at most $T \times L$ entities. Simultaneous to reasoning, P-INT can trace all paths from the head entity of the query to every extended entity in the subgraph. In the matching component, P-INT calculates the similarity between every two paths. Then the RBF aggregation function in \cite{xiong2017deeppath} is used to extract similarity features of the similarity matrix as the interaction between paths. Inspired by FAAN~\cite{sheng2020adaptive}, P-INT computes relation-aware attentions for different paths to model their impact on the matching result based on the relevance of a path to the query relation. 

\subsection{Ontology-based Methods} \label{sec:ontology-based}
Apart from the FSL methods that are mentioned above, there is an emerging interest in extracting knowledge from large language models as pre-training/transforming fine-tuning models have become a default paradigm for natural language processing. Thus, how to effectively transfer between structured relational knowledge and natural language knowledge has become a challenging task. Recent works attempt to integrate structural knowledge like ontology to enhance language understanding. One of the representatives on this track is OntoZSL~\cite{geng2021ontozsl}, which proposes a novel zero-shot learning framework that not only enhances the class semantics with an ontological schema. It also employs an ontology-based generative adversarial network (GAN), as in ZSGAN, to synthesize training samples for unseen classes. OntoZSL first designs an ontology encoder for learning relation representations from the ontological schema by considering the structural relation between concepts and their correlations in the textual description. Pre-trained TransE is used to generate the default embedding of all concepts in the ontological schema. Then a text-aware semantic embedding model is employed by projecting the structural and textual representations into the same embedding space and learning them simultaneously with the same scoring function as in TransE. With two types of representation learned with the ontology encoder, OntoZSL then follows GAN~\cite{qin2020generative} to learn and train the real relation embeddings in bags containing all the one-hop neighbor triples of the task relation. The embeddings of all entity pairs in the bag consist of the real embedding of the task relation, which contains semantic and structural features of the task relation. OntoZSL finally generates plausible relation embedding for each unseen relation with its text description using the well-trained generator. For prediction, the model calculates the similarity between the relation embedding and the candidate's head, tail entity pair. 

\subsection{FKGC with Commonsense Knowledge} \label{sec:FKGC-commonsense}
Previous models primarily focus on structural information in KGs like one-hop neighbors and paths but relatively ignore semantic information. Models like ADK-KG~\cite{zhang2022hg-adk-kg} consider content information but still rely on neighbors to generate embeddings for entities and relations. As mentioned, due to the recent broad investigation of pre-trained language models such as BERT~\cite{devlin2018bert} and GPT~\cite{brown2020language}, some methods are developed by fine-tuning these models to exploit textual information for few/zero-shot KG completion. This section covers several representatives in this category to discuss how to leverage commonsense/textual information in KGs effectively. 

ConMask~\cite{shi2018open} is one of the first models that is proposed to tackle the zero-shot KGC problem by encoding unseen entities with their names and text descriptions. In general, it feeds the text embeddings of the entities and the relation of a triple into a model composed of an attention-based relation-dependent text masking module and a CNN-based target fusion module. To capture text information that is relevant to task relation, ConMask uses a relation-dependent content masking module to reduce noise in the given descriptions. The component first pre-processes the input description to select small relevant snips based on the task relation. ConMask utilizes the attention mechanism to mask the irrelevant task, which assigns a relation-dependent similarity score to words. A common problem here is that the words with the highest scores are not the target entity but words with similar semantic meaning to the task relation. Since actual target words are always around these indicator words, ConMask adjusts the similarity score of each word based on its context. Then the model extracts relation-based entity embeddings via target fusion. Three fully convolutional neural networks (FCN) layers without de-convolution
operations are developed for this task. Since directly generating entity embeddings with the target fusion module may be costly, ConMask employs a semantic averaging function that aggregates word embeddings to represent entity names and generates representations of other textual features for each entity. 

Although ConMask successfully learns embeddings of the entity's name and parts of its text description to connect unseen entities to the KGs, it does not take full advantage of the rich feature information in the text descriptions. Besides, the proposed relationship-dependent content masking method in ConMask may quickly fail to find the target words. To challenge such problems, a Multiple Interaction Attention (MIA) model~\cite{niu2021open} is proposed to acquire the interactions between the head entity description, head entity name, the relationship name, and the candidate tail entity descriptions for more graphic representations. Besides, MIA similarly uses the additional textual features of head entity descriptions to enhance the head entity representation and apply the attention mechanism between candidate tail entities to enhance their representation of them. Specifically, MIA first transforms each word in the head entity description, question, and candidate tail entity description into several continuous representations, including GloVe, POS, NER, and BERT embeddings, and concatenates them to form the input representations of each word. Then, to enhance entity representations with the interaction with all the relevant textual information, MIA leverages the same word-level sequence alignment attention mechanism for each interaction since words in the same description are not equally important, and relevant descriptions usually mention each other. The third component of the model is an RNN layer which uses Bi-LSTM to encode text context. The exact attention mechanism is applied between multiple candidate tail entity descriptions to enhance their representations. MIA also explores different scoring functions to enhance the convergence of the model, which achieves significant improvements against other state-of-the-art models. However, the problem with this approach is that it relies heavily on entity descriptions and only works when necessary information is available. 

InductiveE~\cite{wang2021inductive} proposes a commonsense KG link prediction method that can deal with unseen entities by utilizing textual entity descriptions. It enables inductive learning by directly building representations from entity descriptions instead of leveraging textual entity representations as training initialization. Specifically, It first represents an entity using the concatenation of its text embeddings by the fastText word embedding model~\cite{bojanowski2017enriching} and the last layer for $[CLS]$ token of the pre-trained BERT~\cite{devlin2018bert}. To further enhance semantic entity representations with entity neighbor information, InductiveE adds similarity links for unseen entities to initialize their neighbor information. It then feeds the entity representations of the densified graph into a model composed of a gated-relational GCN encoder and a simplified ConvE~\cite{dettmers2018convolutional} decoder to predict each triple's score. The gated encoder is employed to guarantee adaptively control over the amount of information fused to the center node from their neighboring connections. For the decoder, Conv-TransE~\cite{shang2019end} has been proven effective and efficient in scoring triples in KGC. The decoder of InducctiveE further improved it by adding a shuffling operation before convolution to allow more interaction between embeddings and improve the convolutional model's expressive ability. 

InductiveE only explores inductive learning on unseen entities, while inductive learning on unseen/new relations is also valuable for real-world commonsense FKGC tasks. A KGC model, KG-BERT~\cite{yao2019kg}, is developed to target such a challenge. It transforms a triple head entity, relation, and tail entity into a text sequence. It then makes triple prediction as a downstream text classification task, where BERT is fine-tuned with given training triples. For unseen entities and relations with name information, the candidate triples associated with them can be directly predicted by transforming them into text sequences. 

Following this approach, BERTRL~\cite{zha2022inductive} also proposes to predict triples as a downstream text classification task of BERT, utilizing the text information of entities and relations. However, it fine-tunes BERT using single triples and possible paths connecting two entities where reasoning is conducted explicitly. Given a query triple $(h, ?, t)$ to exploit the neighborhood knowledge of head and tail entities and select proper neighbors for efficiency concern, BERTRL entirely relies on BERT to encode such information. To formalize structural knowledge in KGs to fit into BERT models, BERTRL collects all the length-$k$ reasoning paths between the head and tail entity in the query triple. It then takes each path as a separate input to BERT. Each path individually induces the query triple with a confidence score. The problem is then transformed into a binary classification problem where the score of the linear layer on top of $[cls]$ indicates the correctness of the query triple given a reasoning path. The maximum aggregation of bag scoring is used at inference time to generalize the score of all reasoning paths. The path with the highest score can be used to explain the reasoning process of the prediction. 

As all these works manage to consider textual information simultaneously with structural information, they still treat them as two types of knowledge. On the other hand, prompt-tuning proposed in GPT3~\cite{brown2020language} as an arising methodology has been used for tasks like relation extraction names, entity recognition, etc. Recent works have tried to integrate external knowledge into prompt designing. OntoPrompt~\cite{ye2022ontology} is one of the representatives of this approach. It utilizes prompts to bridge commonsense knowledge from pre-trained language models (LMs) and structural knowledge from knowledge graphs for the FKGC task. OntoPrompt first employs ontology transformation to enrich and convert structure knowledge to text format, where it utilizes pre-defined templates to convert knowledge to text as prompts. Specifically for KGC, the model leverages head entity types and tail entity types from the ontology representation as constraints. 
It uses corresponding items obtained from the external Wikidata query as the source of ontology and extracts the textual description. It follows KG-BERT to consider KGC as a triple classification task and concatenate entities and relations as an input sequence. It also uses the learnable virtual token to enhance the prompt representation. 
Next, OntoPrompt proposes span-sensitive knowledge injection to select informative knowledge. Considering that irrelevant and noisy knowledge may lead to changes in the meaning of the original sentence, OntoPrompt leverages a visible matrix based on spans to limit the impact of corresponding knowledge on the knowledge injection. In this way, not all tokens in the input sentences will be affected by external knowledge. Third, OntoPrompt develops a collective training algorithm to optimize representations jointly. Note that the injected external knowledge should be associated with the surrounding context; learnable tokens are added with random initialization and optimized along with injected ontology tokens with a fixed language model. Inspired by the previous study~\cite{gu2022ppt} that prompt-tuning in the low-data regime is unstable and may obtain poor performance, the model further optimizes all parameters to train the ontology text and input text representations collectively. With the components above, OntoPrompt can enrich task-relevant knowledge using pre-trained LMs, prevent negative knowledge fusion, and integrate commonsense knowledge into structural which solve challenging problems in knowledge missing, knowledge noise, and knowledge heterogeneity and achieve promising performance in FKGC task. 

ZS-SKA~\cite{gong2021prompt} also utilizes prompt to tackle the FKGC task. Instead of leveraging ontology knowledge, they directly work on semantic knowledge augmentation for zero-shot relation classification.

ZS-SKA first generates augmented instances with unseen relations from instances with seen relations following a word-level sentence translation rule. To encode every instance, ZS-SKA tokenizes all the words in a sentence and feeds them to BERT to generate a contextual representation for each token. A CNN layer is used after obtaining the tokenized input sentence. Then, they design prompts based on ConceptNet~\cite{speer2013conceptnet} to integrate semantic knowledge information learned from seen relations and exploit such knowledge to infer the features of unseen relations. They consider multiple types of semantic knowledge, including relation descriptions and name entities, to learn unseen relations effectively. The input sequence is wrapped with a natural language snip template. Instead of using the actual label sets in the prompt template, they automatically construct weighted virtual label words based on the knowledge graph of each label. Prompts are represented by embedding the super-class of the input words and the virtual label embedding for unseen relation. By generating the representations of both seen and unseen relations with augmented instances and prompts through prototypical networks~\cite{snell2017prototypical}, distance is calculated to predict unseen relations.

The works above have already shown how pre-trained language models, as external sources, can help with FKGC tasks for both entity and relation learning. Taking one step further, instead of completing a zero/few-shot query tuple, embraced by the power of pre-trained LMs, another trending topic is to directly complete KGs by constructing triples for unseen entities. Representative models on this task include COMET~\cite{bosselut2019comet} and BERTNet~\cite{hao2022bertnet}. 

COMET~\cite{bosselut2019comet} is the first comprehensive study on automatic commonsense knowledge graph completion. It is a generative transformer model over commonsense knowledge which learns to generate detailed and diverse commonsense descriptions in natural language. As mentioned in Sec~\ref{sec:Commonsense KG}, one main challenge on commonsense knowledge graph reasoning is that commonsense knowledge, represented by open-text, usually does not fit into a fixed schema. Generally, COMET tackles this problem by constructing commonsense KG/KB by training a transformer over existing tuples as a seed set of knowledge to learn an adaptive representation of commonsense knowledge with a pre-trained LM. Then the LM can be used to produce novel tuples with unseen entities. The relations are identified by generating phrases that can semantically complete an existing seed phrase and relation type. In detail, the transformer language model in COMET follows the structure of GPT~\cite{radford2018improving}, which consists of multiple transformer blocks of multi-headed attention and fully connected layers to encode input text. The input of the model is concatenated sequence of words for knowledge tuples. To encode the order information between tokens that is ignored by the transformer, COMET adds a position embedding for each position in the sequence. The position embedding and word embedding of each word is added for the final representation, which is the input to the first transformer layer. COMET is trained to produce the tail entity given the tuple's head entity and relation, which follows the setting of the KGC task but is expected to generate novel tuples that do not exist in the training set. 

The limitation of COMET is that it can only generate triples for new entities with seen relations. The ideal zero-shot KGC model should be able to construct tuples with unseen entities and relations. BERTNet~\cite{hao2022bertnet} is then proposed to harvest KGs with implicit knowledge from pre-trained LMs. BERTNet only requires a few-shot seed set, including an initial prompt and seed entities for each relation as input, and can extract knowledge for unseen relations. In general, the model automatically generates different prompts and searches within a given LM for novel knowledge. 

BERTNet tackles two challenges in KGC/KG construction. First, LMs have shown to be inconsistent even with a slightly different prompt, making it difficult to extract knowledge reliably from LMs. An intuitive solution is to learn the optimal prompts automatically. Such methods require extensive training data, which is unavailable in few-shot or zero-shot settings. To this end, BERTNet employs unsupervised paraphrasing on an initial prompt to generate a set of various prompts with their confidence. Then entity pairs that consistently satisfy these prompts are extracted to generate novel triples. Another challenge then comes into space when searching for proper entity pairs due to the ample candidate space in LMs. BERTNet devises an efficient search-and-rescoring strategy that strikes the balance between knowledge accuracy and coverage.
a prompt/entity pair compatibility function is designed to dynamically reassign weights for both candidate prompts and entities at each knowledge searching step. Specifically for entity searching, BERTNet first uses individual compatible score, which is more accessible to threshold and prune, to weighted average across all prompts to generate a large set of candidate entity pairs. These candidates are then re-ranked by the total compatible score to select the output entity pairs. 

Since BERTNet only requires a set of seed prompts and few-shot entities for relations other than a pre-trained LM, the model guarantees the flexibility to extract novel knowledge even for relations that have complex structures or include multiple entities. Besides, the resulting triples can be considered as an interpretation of the respective black-box LMs. Another novelty of BERTNet is that instead of only looking at matrices like hits@10 and BLUE score, it directly integrates the generated tuples into background KGs and applies the new KGs for downstream tasks. The performance on those tasks indicates that BERTNet can generate novel high-quality tuples. 

\section{Applications and Resources} 
\subsection{Applications} \label{sec:application}
FKGC models have been applied to problems other than prediction tasks on knowledge graphs. One main application is to leverage the few-shot link prediction technique for other graph-related tasks. For example, Meta-Graph~\cite{bose2019meta}investigates few-shot link prediction on different networks, including protein-protein interaction (PPI) networks~\cite{zitnik2017predicting}, 3D point cloud data~\cite{neumann2013graph} and academic social networks~\cite{tang2008arnetminer}. MetaTNE~\cite{lan2020node} also leverages a meta transformer commonly used in FKGC tasks to predict protein-protein interaction. Similarly, molecular property prediction has always been a demanding problem since manual prediction can be costly and inefficient. Meta-MGNN~\cite{guo2021few}and Pre-PAR~\cite{wang2021property} have been proposed to solve such tasks with few-shot link prediction methods. Meta-MGNN takes each molecule as a graph and uses a graph-level-GNN to encode each molecule. It further introduces an attention mechanism to make MAML aware of molecular property differences to improve model encoding further. Moreover, Pre-PAR improves Meta-MGNN by capturing relational structure among different molecular properties to effectively and efficiently propagate limited labels among similar molecules. Similarly, GEN~\cite{baek2020learning} has also been used on drug-drug interaction prediction. 

Another challenging real-world task related to link prediction is recommendation problems. For example, as proposed in the Yelp challenge~\cite{asghar2016yelp}, user reviews have become a significant part of web services like Yelp. Since users can post their opinions about businesses, products, and services through reviews consisting of free-form text and a numeric star rating, the interaction between users, services/products, and reviews intuitively form structural and textual knowledge. Recent techniques in collecting user-related information, like GPS-enabled devices, help form location-based social networks that provide the location information that is valuable for the recommendation system. One challenge for such systems is that when a new user joins, there is little existing knowledge in the platform besides basic information and location. Thus, few-shot link prediction models like SEATLE~\cite{li2020few} and heterogeneous information network-based models~\citep{zhao2017meta,lu2020meta} aim to tackle cold-start recommendation problems over graphs with meta-learning. Other models~\cite{yang2022few-1, ding2020graph, wang2020graph, liu2021relative, yao2020graph, tan2021graph, ding2021few} tackles recommendation problems with methods like attribute matching with previous users, transforming knowledge from other platforms with cross-network meta-learning, modeling user with updated information with dynamic meta-graph reasoning, etc. 

As mentioned in BERTNet~\cite{hao2022bertnet}, since FKGC models help to improve the quality and coverage of original KGs, the output can be leveraged by downstream tasks. For example,  affordance reasoning and extraction~\cite{zhu2014reasoning} can used few-shot KGC models to generate affordance for unseen entities with pre-trained LMs or similarity-matching with seen entities in the background KG. More recently, in large-scale video extraction/recognition datasets like EPIC~\cite{damen2018scaling} and STAR~\cite{wu2021star} or action planning tasks, procedural reasoning is required due to the rapid changing of reasoning scenarios and dynamic environment knowledge. FKGC models can be powerful on such tasks since they can easily capture new few-shot information and swiftly adapt and optimize themselves to fit each reasoning step. 

\begin{table*}
 \caption{Statistics of FKGC Benchmarks}
 \label{tab:FKGC benchmarks}
    \begin{subtable}{\textwidth}
        \caption{Relation prediction Benchmarks}
        \centering
        \begin{tabular}{cccccc}
             \toprule
             & Ent \# & Rel \# & Triples \# & Task rel \# (train/valid/test) & Source \\
             \midrule
             NELL-One~\cite{xiong2018one} & 68,545 & 358 & 181,109 & 51/5/11 & NELL \\
             \midrule
             Wiki-One~\cite{xiong2018one} & 4,838,244 & 822 & 5,859,240 & 133/16/34 & Wikidata \\
             \midrule
             NELL-ZS~\cite{qin2020generative} & 65,567 & 181 & 181,109 & 139/10/32 & NELL \\
             \midrule
             Wiki-ZS~\cite{qin2020generative} & 605,812 & 537 & 724,967 & 469/20/48 & Wikidata \\
             \bottomrule
        \end{tabular}
    \vspace{0.5cm}
        \caption{Entity prediction Benchmarks}
        \centering
        \begin{tabular}{cccccc}
             \toprule
             & Ent \# & Rel \# & Triples(train/valid/test) \# & Unseen Ent \# (train/valid/test) \# & Source \\
             \midrule
             WN18RR-sub~\cite{baek2020learning} & 4,478 & 11 & 93,003 (-) & 3000 & WN188RR \\
             \midrule
             FB15K-237-sub~\cite{baek2020learning} & 10,938 & 237 & 72,065/6,246/9,867 & 2,500/1,000/1,500 & FB15K-237 \\
             \midrule
             NELL-995-sub~\cite{baek2020learning} & 5,694 & 200 & 22,345/3,676/5,852 & 1,500/600/900 & NELL-995 \\
             \midrule
             FB15K-237-OWE~\cite{shah2019open} & 14,405 & 235 & 242,489/10,963/36,250 & 2081 (-) & FB15K-237 \\
             \midrule
             Wikidata5M~\cite{wang2021kepler} & 4,594,485 & 1222 & 20,496,514v/6,699/6,894 & 4,579,609/7,374/7,475 & Wikidata \\
             \bottomrule
        \end{tabular}
    \end{subtable}
 
\end{table*}

\subsection{Resources} \label{sec:resources}
With the bursting growth of zero-shot KG completion methods, various benchmarks have been proposed for evaluation. They are usually constructed based on existing commonly used typical KG completion datasets, including FB15k~\cite{bollacker2008freebase}, FB15k-237~\cite{pellissier2016freebase}, NELL-995~\cite{xiong2017deeppath}, and some other sub-KGs extracted from popular KGs such as Wikidata~\cite{vrandevcic2014wikidata}. Such benchmarks usually obtain entity information from the original KGs. For example, DBpedia50k, FB20k, and Wikidata5M collect correspondingly text descriptions of entities from DBpedia and Wikipedia. Here we introduce several representatives for zero/few-shot KGC:
\begin{itemize}
    \item \textbf{FB15k-237-OWE}~\cite{shah2019open} is built on FB15k-237 for zero-shot KGC. They first sample a set of tail entities and randomly pick associated head entities from FB15K-237. Then all triples with their head in the associated entity set are moved to the testing set, which forms the testing set for tail entity prediction. At the same time, the ones with their tail in the associated entity set are removed. The testing set for head entity prediction is similarly generated. These two sets form the final testing set by further removing triples whose relations are not included in the training set. This testing set is further split into a validation set and the final testing set. The dataset contains 2,081 unseen entities, 12,324 seen entities, and 235 relations. The numbers of triples for training/validation/testing are 242,489/10,963/36,250. \\
    
    \item \textbf{Wikidata5M}~\cite{wang2021kepler} was originally constructed for evaluating text-aware KGE models but is now widely used for zero-shot KGC. It is developed based on Wikidata and English Wikipedia dump. Each entity uses the first section of its Wikipedia page as its description. The dataset excludes entities without Wikipedia pages or descriptions shorter than 5 words. Next, they extract all the triples from Wikidatadump. The dataset keeps triples with qualified head and tail entities, leaving it with 4,594,485 entities, 822 relations, and 20,624,575 triplets. To support the zero-shot setting, they randomly extract two sub-KGs as the validation and testing sets and use the remaining as the training sets. They ensure that the entities and triples are mutually disjoint across the training, validation, and testing sets. Detailed information on each set can be found in Table~\ref{tab:FKGC benchmarks}. \\
    
    \item \textbf{Subsets of WN18RR, FB15k-237, and NELL-995} are constructed by \cite{baek2020learning} for out-of-graph completion between seen, unseen entities, and unseen entities themselves. These subsets are systematically extracted from their original benchmarks. They randomly sample a group of entities with less than 100 associated triples as unseen entities. These triples are further separated to compose three meta sets (meta-training, meta-validation, and meta-testing sets). The rest of the original benchmarks are considered as the background KGs/In-Graph, and entities inside are respectively seen entities. The meta sets are then cleaned to guarantee that each of their triples has at least one unseen entity and that all the triples are out of In-Graph. More statistics about seen/unseen entities are presented in Table~\ref{tab:FKGC benchmarks}. \\
    
    \item \textbf{NELL/Wiki-ZS} are two zero-shot KG completion benchmarks. Each benchmark contains three relation-disjoint sets: the training set holds seen relations, validation/testing set holds unseen relations. Associated triples are separated accordingly, while entities included in the testing/validation set are all involved in the training set. NELL-ZS has 139/10/32 relations in the training/validation/testing set
    , while Wiki-ZS involves 469/20/48 relations for training/validation/testing.
    GEN~\cite{qin2020generative} uses relation textual descriptions as the textual information, while OntoZSL~\cite{geng2021ontozsl} constructs ontological schemas, which contain not only textual information but also other relation knowledge, including relation hierarchies and relation domains for both benchmarks. \\ 
    
  \item \textbf{NELL/Wiki-One} are originally developed in GMatching~\cite{xiong2018one} for evaluating few-shot KG completion with unseen relations only one supporting instance. 
  To construct the testing set, they extract relations with less than 500 but more than 50 associated triples from the original benchmarks and use those as task relations. With this preprocessing, 67 relations are extracted in NELL-One, and the benchmark further partitions them into 51/5/11 to further extracted associated triples and compose the training/validation/testing set. Similarly, in Wiki-One, 183 relations are extracted and partitioned into 133/16/34 for constructing triples in the training/validation/testing set. On the entity side, 68,545 entities are extracted for NELL-One, and 4,838,244 for Wiki-One. Additionally, another 291 and 639 relations are extracted as background relations to construct more triples for the entities. The relations in the training/validation/testing set are guaranteed to be disjoint so that the two benchmarks can be used for both zero-shot KGC and few-shot KGC tasks. 
  
\end{itemize}

\section{Analysis}
Starting with GMatching~\cite{xiong2018one}, since the problem setting comes from KGC problem, FKGC models intuitively enhance KGE or GNN models which originally capture the structural information with meta-learning frameworks. Early metric-based models like MetaR, FSRL~\cite{zhang2020few} and FAAN~\cite{sheng2020adaptive}, focusing on how to effectively leverage neighbor information by assigining static or task-aware attentions to different neighbors. Optimization models like Meta-KGR~\cite{lv2019adapting}, ADK-GK~\cite{zhang2022hg-adk-kg}, ZS-GAN~\cite{qin2020generative}, explore to improve entity and relation embeddings with multi-hop path information and. These paths can also be used to explain reasoning logic which improve the  interpretablility of the models. Meta-learning provides the opportunity for these models to quickly adapt to few-shot tasks at testing stage. 

However, these models ignore the semantic information in background KGs like entity/relation names, text description. To combine the two types of information, prompt-based models like OntoPrompt~\cite{ye2022ontology} and ZS-SKA~\cite{gong2021prompt} are proposed. These models employs prompts to translate triples in KGs into sentence-like knowledge. Pre-trained language models like BERT are then used to generate entity and relation embeddings. Semantic information like text description of entities and relations helps to enrich the embeddings. 

All these models are dependent on large-scale KGs since they provide structural information for entity and relation embeddings. On the other hand, GPT models~\cite{brown2020language, radford2018improving} show that pre-trained language models originally contain structural knowledge as well, some models take a step further to totally depend on pre-trained LMs. COMET~\cite{bosselut2019comet} and BERTNet~\cite{hao2022bertnet} are the representatives of this category. COMET follows the framework of GPT models and BERTNet uses prompts to generate novel triples. Comparing with models with structural knowledge, these models can usually work on both structural and commonsense KGs. This approach free FKGC models from background KGs and extend the usage to downstream tasks in more area. 

\begin{table*}[t]
\centering
\resizebox{1\textwidth}{!}{
    \begin{tabular}{ccccccccc}
        \toprule
        & GMatching~\cite{xiong2018one} & FSRL~\cite{zhang2020few} & MetaR~\cite{chen2019meta} & OntoPrompt~\cite{ye2022ontology} & InductiveE~\cite{sheng2020adaptive} & COMET!\cite{bosselut2019comet} & BERTNet~\cite{hao2022bertnet}\\
        \midrule
        NELL-One~\cite{xiong2018one} & \cmark & \cmark & \cmark & \xmark & \xmark & \xmark  & \xmark\\
        FB15K-237-sub~\cite{baek2020learning} & \cmark & \cmark & \cmark & \cmark & \xmark & \xmark  & \xmark\\
        WN18RR-sub~\cite{baek2020learning} & \cmark & \cmark & \cmark & \cmark & \xmark & \xmark  & \xmark\\
        NELL-ZS~\cite{qin2020generative} & \xmark & \xmark & \xmark & \xmark & \xmark & \xmark  & \xmark\\
        \midrule
        ConceptNet~\cite{speer2013conceptnet} & \xmark & \xmark & \xmark & \xmark & \cmark & \cmark  & \cmark\\
        ATOMIC~\cite{sap2019atomic} & \xmark & \xmark & \xmark & \xmark & \cmark & \cmark  & \xmark\\
        \toprule
    \end{tabular}
}
\caption{Evalution of Representative FKGC Models}
\label{tab:FKGC Evaluation}
\end{table*}

\section{Future directions}
As discussed, earlier FKGC models are primarily extensions of meta-learning and transfer learning methods to FKGC. Another type of knowledge in meta-learning is the learning process. In the future, representing knowledge on learning (e.g., previous reasoning process~\cite{suris2020learning}) as KGs and integrating them with meta-learning or transfer learning algorithms could lead to more general neural-symbolic paradigms that apply to different FSL tasks. Propagation-based methods like GEN~\cite{baek2020learning} and REFORM~\cite{wang2021reform} solve few-shot KG completion by utilizing the few-shot samples, i.e., triples model the correlation of unseen entities with the distribution and correlation of seen ones. It would be a promising solution to utilize these few-shot links and the unseen entities' correlations auxiliary information such as textual descriptions, attributes, and schemas.

Another track of FKGC models like OntoPrompt~\cite{ye2022ontology} has proved that exploiting ontology/rule structured knowledge is a promising approach to infer symbolic knowledge like triples for unseen entities/relations. On the other hand, generation-based models like OntoZSL~\cite{geng2021ontozsl} are not biased to seen or unseen classes in prediction compared with the widely explored mapping-based methods~\cite{chen2021low}. Therefore, generation-based ZSL methods conditioned on the embeddings of KGs could be a future direction for FKGC task.

It is till recently that semantic information has gained attention in FKGC models. GANA~\cite{niu2021relational}first proposes to integrate the semantics of a few-shot relation's neighborhood, and ZSGAN~\cite{qin2020generative} generates reasonable relation embeddings with text representations of task relations. Works like COMET~\cite{bosselut2019comet}, KG-BERT~\cite{yao2019kg}, and BERTNet~\cite{hao2022bertnet} further present the effectiveness of learning for representation of unseen/few-shot entities and relation. The performance of these models indicates that multi-modal knowledge can also be beneficial for FKGC tasks. Challenging the problem of efficiently integrating data from more modalities into current FKGC models can be a promising direction to tackle not only knowledge graph completion tasks but also zero/few-shot reasoning in other areas. 

According to Table~\ref{tab:FKGC Evaluation}, currently there is not enough evaluation to compare the performance between FKGC models with large-scale KGs and the ones leveraging pre-trained LMs. It is still at theoretic level that models like BERTNet and COMET are effective for strucural KGs like NELL and Freebase. Such evaluations will also be valuable for few-shot knowledge graph completion with multi-modal information. 

\section{Conclusion}
As knowledge graphs are a popular source for tasks in various domains, few-shot knowledge graph completion models provide a chance to efficiently integrate new knowledge into existing KGs to improve the quality and coverage of KGs. In this survey, we first introduce the major challenges and bases of FKGC. Then we comprehensively reviewed previous studies on FKGC. We categorize these methods into two groups: FKGC with structural knowledge and FKGC with semantic knowledge. This categorization shows the trending of FKGC methods from meta-learning and attention-based models to leveraging semantic and textual information or even directly extracting structural triples from pre-trained language models. Then we discuss how FKGC models can be transferred or applied in various fields. Finally, we summarize the remaining challenges for different FKGC models and possible future directions. We hope this survey will serve as a valuable guide for others who are interested in few-shot knowledge graph completion and advance future works in this area. 

\bibliographystyle{ACM-Reference-Format}
\bibliography{ref}

\appendix

\end{document}